\documentclass[final]{cvpr}

\usepackage{times}
\usepackage{epsfig}
\usepackage{graphicx}
\usepackage{amsmath}
\usepackage{amssymb}
\usepackage{xr}
\usepackage[pagebackref,breaklinks,colorlinks]{hyperref}

\graphicspath{{./figs/}{./figs/result/}}
\title{A Plug-and-Play Defensive Perturbation for Copyright Protection of DNN-based Applications}

\author{Donghua Wang\\
Zhejiang University\\
\and
Wen Yao, Tingsong Jiang, Weien Zhou, Xiaoqian Chen\\
Chinese Academy of Military Science\\
\and
Lang Lin\\
Zhejiang Economic Information Center\\
}

\begin{document}

\maketitle

\begin{abstract}
Wide deployment of deep neural networks (DNNs) based applications (e.g., style transfer, cartoonish), stimulating the requirement of copyright protection of such application's production. Although some traditional visible copyright techniques are available, they would introduce undesired traces and result in a poor user experience. In this paper, we propose a novel plug-and-play invisible copyright protection method based on defensive perturbation for DNN-based applications (i.e., style transfer). Rather than apply the perturbation to attack the DNNs model, we explore the potential utilization of perturbation in copyright protection. Specifically, we project the copyright information to the defensive perturbation with the designed copyright encoder, which is added to the image to be protected. Then, we extract the copyright information from the encoded copyrighted image with the devised copyright decoder. Furthermore, we use a robustness module to strengthen the decoding capability of the decoder toward images with various distortions (e.g., JPEG compression), which may be occurred when the user posts the image on social media. To ensure the image quality of encoded images and decoded copyright images, a loss function was elaborately devised. Objective and subjective experiment results demonstrate the effectiveness of the proposed method. We have also conducted physical world tests on social media (i.e., Wechat and Twitter) by posting encoded copyright images. The results show that the copyright information in the encoded image saved from social media can still be correctly extracted.
\end{abstract}

\begin{figure}
	\centering
	\begin{minipage}{.24\linewidth}
		\centering
		\includegraphics[width =1\linewidth]{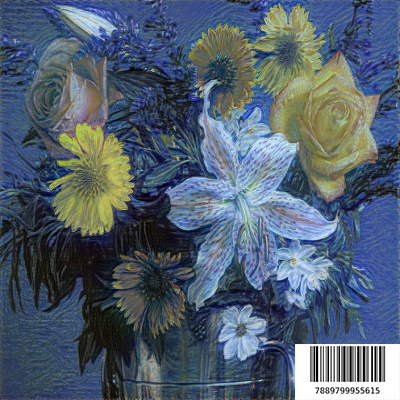}
		\centerline{\footnotesize (a) Barcode}
	\end{minipage}
	\begin{minipage}{.24\linewidth}
		\centering
		\includegraphics[width =1\linewidth]{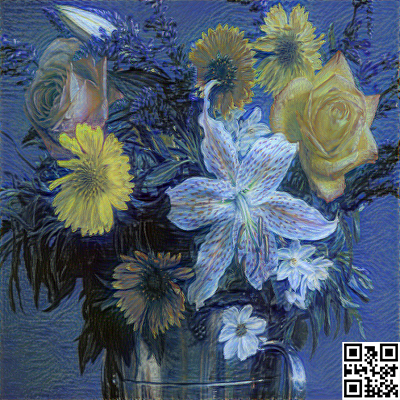}
		\centerline{\footnotesize (b) QR Code}
	\end{minipage}
	\begin{minipage}{.24\linewidth}
		\centering
		\includegraphics[width =1\linewidth]{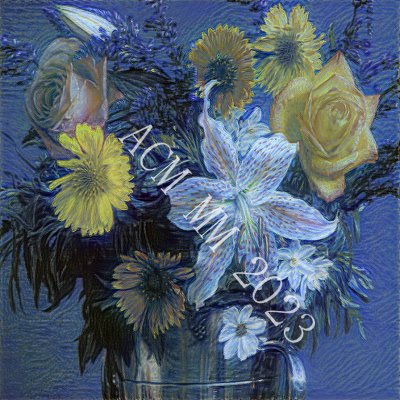}
		\centerline{\footnotesize (c) Visible Watermark}
	\end{minipage}
	\begin{minipage}{.24\linewidth}
		\centering
		\includegraphics[width =1\linewidth]{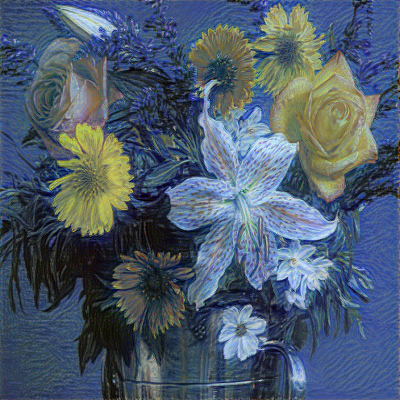}
		\centerline{\footnotesize (d) Our method}
	\end{minipage}
	\caption{Common techniques used for copyright protection (a)-(c) and the technique proposed by our proposed method (d). Obviously, the image with the past copyright protection techniques is visually conspicuous, and the completeness of the content has been impaired. In contrast, the image with our method is inconspicuous, featuring excellent stealthiness.}
	\label{fig:teaser}
\end{figure}

\section{Introduction}
Recently, the large-scale deep learning model has made a revolutionary development in engineering implementation, emerging many phenomenally applications, such as ChatGPT and DALL$\cdot$E 2, imposing a prudent influent on people's daily life. Meanwhile, people can exploit deep learning techniques to create various content(e.g., image style transfer, image cartoonish), such techniques dubbed AI Generated Content (AIGC). In such backgrounds, the copyright protection of AIGC-based commercial applications is becoming urgent and inevitable. Furthermore, given that anyone can create an AIGC-based application as long as he masters the AIGC technique, it is necessary to develop an efficient copyright protection method for AIGC-based commercial applications, such as the deep learning model and its products. 

Currently, some common approaches can be used for copyright protection, including but not limited to registered copyright, copyright statement, and encrypted protection. Such approaches are effective for real entity products (e.g., books, films) in real life while unsuitable for the products yielded by DNN-based applications. Although some traditional copyright protection techniques are available, such as the Barcode and QR code as well as watermark (visible and invisible), they introduce visible traces on the image to be protected (see Figure \ref{fig:teaser}(a)-(c)), impairing the completeness of the image content and then resulting in poor user experience, which is unacceptable in high image quality requirements scenarios. Moreover, the cost and efficiency of traditional copyright protection techniques may be unsuitable for DNN-based applications features in the high visitor volume.

Unlike the aforementioned approaches, we get inspiration from adversarial examples, a kind of example constructed by carefully crafted malicious perturbations that are invisible to human observers but can make DNNs output wrong results \cite{szegedy2014intriguing}. Rather than apply the adversarial perturbation to perform adversarial attacks, some research \cite{salman2021unadversarial,wang2022defensive} explored the feasibility of applying adversarial perturbations to auxiliary DNN's decisions, obtaining surprising performance. Motivated by that, we investigate the feasibility of adversarial perturbation in the copyright protection scenario, wherein the adversarial perturbation is regarded as the carrier of copyright information, which is added to the image to be protected and engender the copyrighted ($\copyright$)image, where the $\copyright$image is imperceptible to human beings but can be decoded by the copyright decoder.

\begin{figure}[t]
  \centering
  	\begin{minipage}{.4\linewidth}
  		\centering
  		\includegraphics[width =1\linewidth]{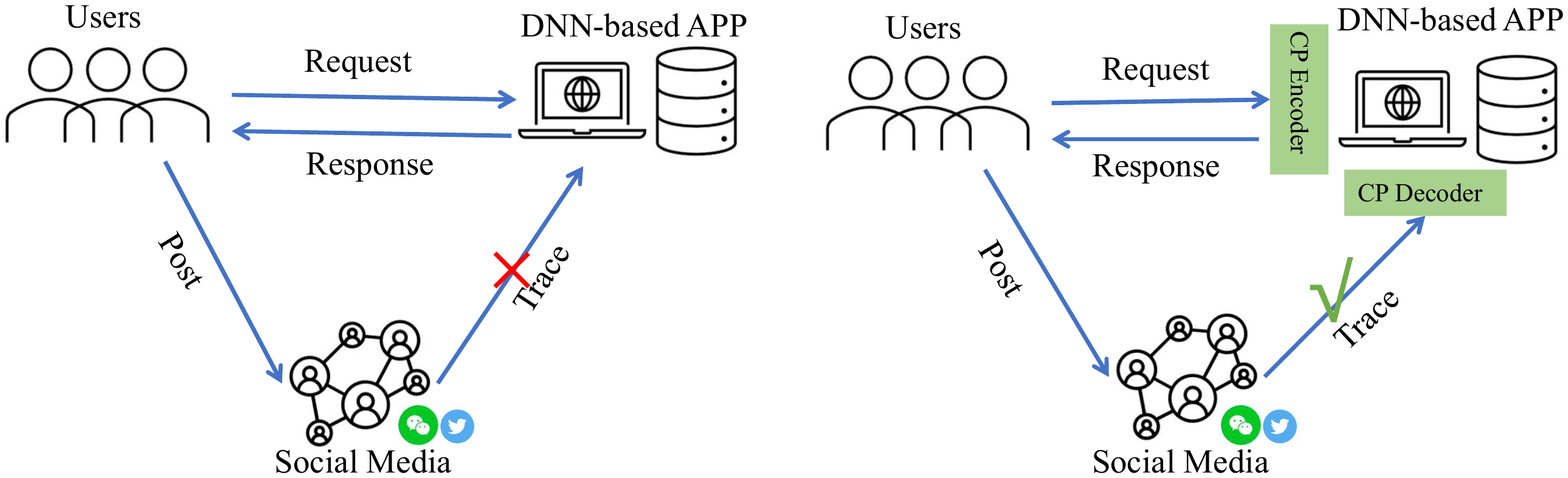}
		\centerline{\footnotesize (a)}
  	\end{minipage}
  	\begin{minipage}{.4\linewidth}
  		\centering
  		\includegraphics[width =1\linewidth]{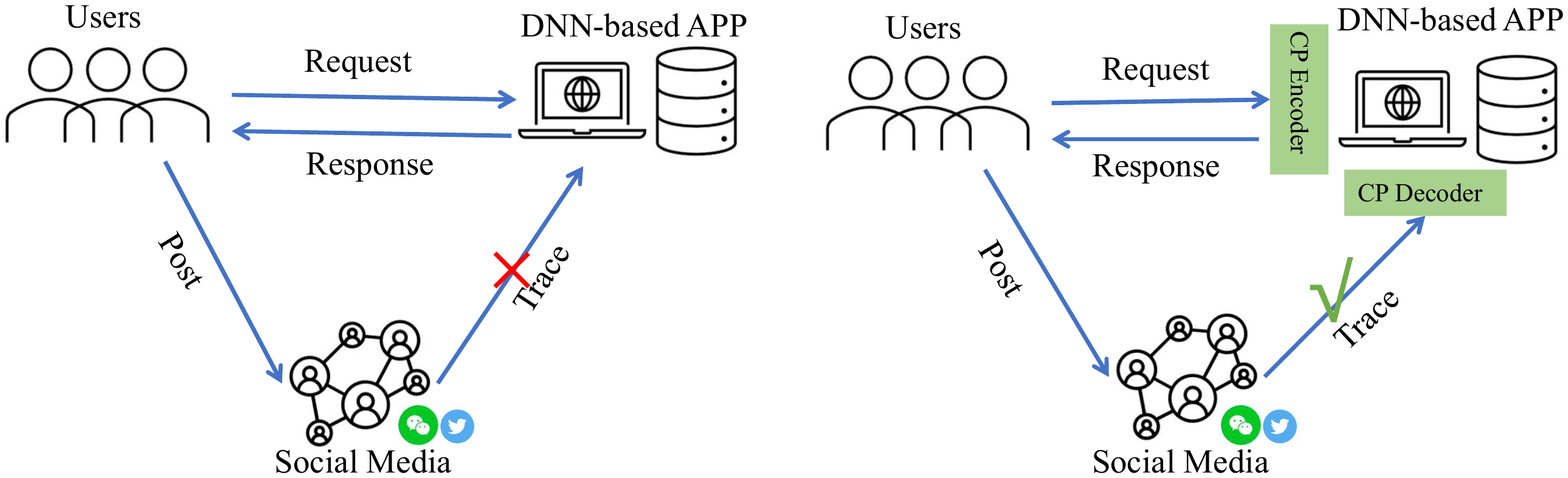}
		\centerline{\footnotesize (b)}
  	\end{minipage}
  \caption{Overview of the DNN-based application (APP) with (a)/without (b) copyright protection.}
  \label{fig:overview}
\end{figure}

In this paper, we proposed a novel plug-and-play invisible copyright protection method based on defensive perturbation for DNN-based applications. We take the style transfer application as an example in our work. Note that the clean image and the styled image have the same meaning, which refers to the image being protected. Specifically, we devise a copyright encoder to project the copyright image to defensive perturbation, which is added to the clean image and engenders the $\copyright$ image. Then, we design a copyright decoder to extract the copyright image from the $\copyright$image. Moreover, we take the potential distortion (e.g., compression) that occurs when the user posts the image to the social media platform into account and devise a robustness module to enhance the anti-disturb capability of the copyright decoder. As an invisible copyright protection method,  the goal is to guarantee the completeness of the decoded copyright image without sacrificing the image quality of $\copyright$image as visible distortion is unacceptable in high image quality requirement scenarios. Thus, a loss function is elaborately devised to ensure the image quality of the $\copyright$image as well as the decoded copyright image. We conduct extensive objective and subjective experiments to demonstrate the effectiveness of the proposed method. Additionally, a physical experiment is performed by posting the $\copyright$image to social media (e.g., WeChat and Twitter) and save for decoding, and the results verify the practicability of the proposed method in the real world. In real scenarios, the service providers can use our method to protect their AIGC products, and they can trace and forensics efficiently with our method if needed, see visualization explanation in Figure\ref{fig:overview}.

Our main contributions are listed as follows.
\begin{itemize}
\item We propose a novel plug-and-play defensive perturbation copyright protection method for DNN-based applications.
\item We devise an encoder-decoder network to realize copyright protection, where the encoder aims to generate invisible defensive perturbation with copyright information while the decoder enhanced by a robustness module is tailored to extract the copyright information from the encoded image.
\item We conduct objective and subjective as well as physical experiments, and the results demonstrate the effectiveness of the proposed method.
\end{itemize}

\section{Related Works}
In this section, we briefly review the DNN-based applications and then discuss the research combined with adversarial examples and watermarks in detail.

\subsection{DNN-based Application}
A line of deep learning techniques has been widely applied in various applications. For example, image-to-image-based techniques have been used in style transfer \cite{gatys2016image,zhu2017unpaired,gu2018arbitrary,park2019gaugan} and cartoonish \cite{chen2018cartoongan,dhir2021automatic} applications (e.g., Prisma), which get an image and style image as input and generate a new image with the given style. In addition, DNN-based face editable techniques, such as the applications of conditional (e.g., editable variable) GANs \cite{choi2018stargan,choi2020stargan,han2021disentangled,yao2021latent,khodadadeh2022latent}, some of them have been embedded in face editing applications (e.g., Adobe Lightroom). Despite DNN-based techniques being widely deployed in various applications, we concentrate on style transfer in this work for simplicity, while our method can easily extend to other applications.

\begin{figure*}[!htbp]
  \centering
  \includegraphics[width=\linewidth]{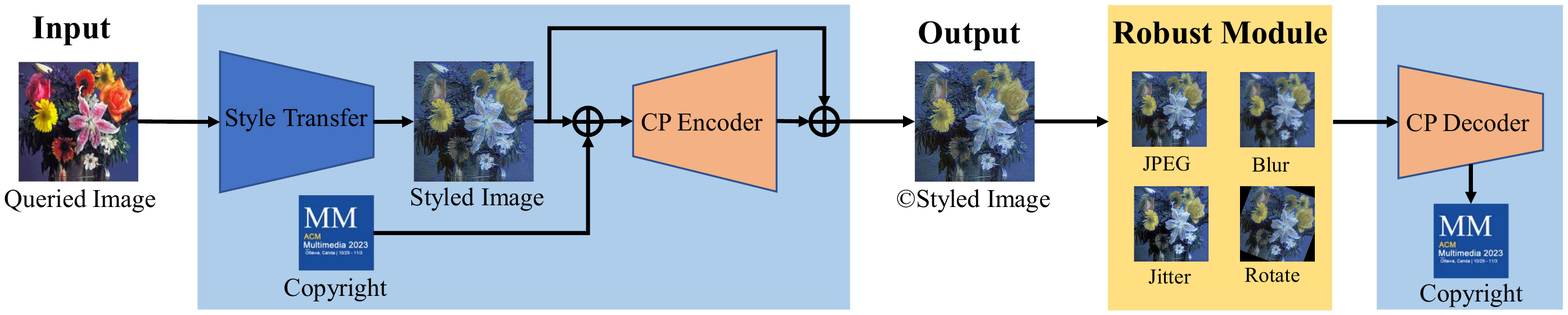}
  \caption{Overview of the proposed method. }
  \label{fig:framework}
\end{figure*}

\subsection{Adversarial examples and watermark}
Adversarial examples attacks are a double-bladed sword. On the one hand, adversarial examples can make the DNNs output incorrect result, imposing potential security risks to security-sensitive applications and may lead to unexpecting loss. For example, the combination of watermark and adversarial examples brings more concrete and stealthy risk. Jia \textit{et al.} \cite{jia2020adv} presented Basin Hopping Evolution algorithm to generate visible adversarial watermarks against the image classification tasks. They treated the common logo (e.g., school badge) as the fixed adversarial pattern but optimized the paste position and the transparency of the logo in the clean image until adversarial examples were generated. Recently, Huang \textit{et al.} \cite{huang2022cmua} proposed a novel approach to generate a universal adversarial watermark against the DeepFake model. The image added with the universal adversarial watermark can mislead the deepfake model to output the undesired fake facial images. 

On the other hand, some researchers attempt to utilize adversarial examples to auxiliary DNNs from a positive perspective. Salman \textit{et al.} \cite{salman2021unadversarial} pointed out that DNN-based applications are easily affected by bad weather environments (e.g., fog and dust) and then are unable to make correct predictions. Thus, the author designed an adversarial texture for a specific 3D object, enhancing the detection accuracy in bad weather. Follow \cite{salman2021unadversarial}, Wang \textit{et al.} \cite{wang2022defensive} concentrated on traffic sign recognition tasks and proposed a method to generate the defensive patch. The traffic sign board pasted with the defensive patch can be correctly predicted by DNNs in the snow environment. 

Currently, the watermark has been adopted to perform adversarial attacks\cite{jia2020adv,huang2022cmua}, while the utilization of adversarial perturbation to protect the DNN-based application remains unexplored, which will be explored in this work.

\section{Methodology}
In this section, we first introduce the problem statement. Then we elaborately introduce the proposed plug-and-play defensive perturbation copyright protection method. Finally, the loss function is described.

\subsection{Problem Statement}
Given a content image ${\rm \textbf{I}}_c$ and style image ${\rm \textbf{I}}_s$, the style transfer method learns a mapping function $F: ({\rm \textbf{I}}_c, {\rm \textbf{I}}_s) \longmapsto {\rm \textbf{I}}$, where ${\rm \textbf{I}}$ is the styled image with the given style ${\rm \textbf{I}}_s$. To realize copyright protection for images generated by style transfer, we devise a copyright (CP) encoder $E$ to engender the defensive perturbation $\delta$ with copyright image $\textbf{I}_{\textit{cp}}$, i.e., $\delta = E(\textbf{I}, \textbf{I}_{\textit{cp}})$. The copyrighted styled image $\copyright \textbf{I}$ obtained by
\begin{equation}
\rm \copyright \textbf{I} = \textbf{I} + \delta.
\label{eq:encoder}
\end{equation}
After get $\rm \copyright \textbf{I}$, a copyright (CP) decoder $D$ is tailored to extract the copyright image $\textbf{I}'_{\textit{cp}}$ from it by
\begin{equation}
\rm \textbf{I}'_{\textit{cp}} = D(\copyright \textbf{I}).
\label{eq:decoder}
\end{equation}
As a plug-and-play auxiliary copyright protection tool, the goal is to ensure the completeness of the decoded copyright image without sacrificing the image quality of the copyrighted styled image $\copyright \textbf{I}$. Therefore, our problem can be modeled as the following form
\begin{equation}
\begin{split}
\rm \min &~~ ||\textbf{I}_{\textit{cp}}- \textbf{I}'_{\textit{cp}}||_p \\
	s.t.&~~ ||\delta||_p \leq \epsilon, 
\end{split}
\label{eq:problem}
\end{equation}
where $||\cdot||_p$ is the $L_p$-norm, which controls the magnitude of the variable, and the $\rm \epsilon$ bound the maximum magnitude of defensive perturbation. We solve the problem defined in Equation (\ref{eq:problem}) by devising a copyright protection network, which will be introduced in detail.

\subsection{Copyright Protection Network}
The copyright protection network comprises two networks: a CP encoder and a CP decoder. CP encoder is devised to engender the defensive perturbation from the combined input of the styled image and copyright image, which is added to the styled image. By contrast, the CP decoder is tailored to extract the output image of the CP decoder. Moreover, a robust module is introduced to enhance the capability of the decoder against the distortion caused by transmission (e.g., compression happens when posting to social media). The overview of the proposed method is illustrated in Figure \ref{fig:framework}.

\subsubsection{Copyright Encoder}
Encoder-decoder is the most common architecture to implement image-to-image tasks. Thus, we adopt the encoder-decoder architecture as our CP encoder network. The previous works (e.g., super-resolution \cite{ledig2017photo,liang2021swinir}) fed the natural image to the encoder-decoder and produced the natural image, which makes the network convergence slow and hard to train. However, we get inspiration from the adversarial attack and propose to use the encoder-decoder to realize a task of image-to-perturbation, which alleviates the learning difficulty of the network. Specifically, we first used a pre-encoder stacked with several convolution layers to map the copyright image to the feature map, which is fused with the input-styled image in a channel-wise concatenate manner and then fed into the CP encoder. The output of the CP encoder is the defensive perturbation with copyright information, which is added to the style image to construct the $\copyright$styled image.

\subsubsection{Copyright Decoder}
The CP decoder requires to decode the copyright image from the encoded $\copyright$styled image. To this end, the CP decoder is an image-to-image task, so we adopt the same architecture (i.e., encoder-decoder) with the CP encoder (except the pre-encoder) as the CP decoder. Note that the output of the CP decoder is the natural image, while the output of the CP encoder is the defensive perturbation. Considering the practical factor, the CP decoder should be equipped with the capability of decoding non-loss images and distorted images. Therefore, the robustness of the CP decoder is significant for its real application in the real world, where the robustness technique will be discussed as follows.

\subsubsection{Robustness Module}
In the real scenario, many factors may impair the encoded $\copyright$styled image and lead to failure in decoding the copyright image. For example, the social media platform (e.g., WeChat, Twitter) may compress the user-posted $\copyright$styled image for commercial factors (e.g., storage cost). The most common compression technique is JPEG compression, which can significantly reduce the desired storage space of images but result in worse image quality. That process may damage the embedded copyright information and make the CP decoder unable to decode the copyright image correctly. Therefore, we adopt the differentiable JPEG \cite{shin2017jpeg} in the optimization process to enhance the anti-disturb ability of the CP decoder. Moreover, we also consider the color jitter and Gaussian blur together to improve the decoding capability further. In addition, we take a step further and use the spatial transformer network (STN) \cite{jaderberg2015spatial}, an effective learnable module used to enhance the network's ability of anti-translation, scale, rotation, and more generic warping. Note that the STN is placed in the CP decoder. Consequently, the decoding process should be reformulated as 

\begin{equation}
\rm \textbf{I}_{\textit{cp}} = \mathbb{E}_{\textit{t}\in \mathcal{T}}D(\textit{t}( \copyright\textbf{I})),
\label{eq:decoder_enhance}
\end{equation}
where $\mathcal{T}$ is the robustness transformation distribution, including different compression quality, blur strengths and so on.

\subsection{Loss Function}
The encode loss is devised to guarantee the copyright image can be embedded into the styled image and achieves the invisible effect. Meanwhile, a decode loss is tailored to ensure the copyright image is correctly decoded from the encoded $\copyright$styled image. To make training more effective, we co-joint training the CP encoder and CP decoder with the following loss function

\begin{equation}
\mathcal{L} = \lambda_1 \mathcal{L}_{enc} + \lambda_2 \mathcal{L}_{dec} + \mathcal{L}_{yuv},
\label{eq:total_loss}
\end{equation}
where $\lambda_1$ and $\lambda_2$ are weights to balance the capability of the CP encoder and CP decoder. $\mathcal{L}_{enc}$ and $\mathcal{L}_{yuv}$ is the loss function to guarantee the image quality of $\copyright$styled image in different feature spaces and benefit in engender more imperceptible defensive perturbation, $\mathcal{L}_{dec}$ is the decode loss used to ensure the image quality of decoded copyright image, which are described as follows.

\textbf{Encode loss function.} Our encode loss function is composed of two-fold from different feature spaces: intermediate feature space of network and image color space. The former aligns encoded $\copyright$styled images and the style images in the intermediate features extracted from a specific network (e.g., VGG16), which is conducive to encouraging the perturbation to concentrate on sematic-related regions, resulting in the better visual quality of encoded $\copyright$styled image. Specifically, we adopt the \textit{LPIPS} loss \cite{zhang2018unreasonable}, a widely used metric that can match human perception well, as the decode loss in intermediate feature space, which is expressed as follows.
\begin{equation}
\mathcal{L}_{enc} = LPIPS(\textbf{I}, \copyright \textbf{I}).
\label{eq:enc}
\end{equation}
Rather than exploit the RGB color space to constrain the magnitude of the defensive perturbation, we adopt the YUV color space as it better coincides with the human visual system. Moreover, some researches \cite{aksoy2020attack,yin2023reversible} shows that hiding information data in YUV color space results in better visual quality. Therefore, we  devise $\mathcal{L}_{yuv}$ in the YUV color space, which is formulated as 
\begin{equation}
\mathcal{L}_{yuv} = ||YUV(\textbf{I}) - YUV(\copyright \textbf{I})||_2.
\label{eq:yuv}
\end{equation}
where $YUV(\cdot)$ denotes the color space convert function that converts an image from RGB color space to YUV color space.

\textbf{Decode loss function.} We observed that the copyright image featured clear structure information in most cases (e.g.,  badge or trademark), encouraging us to adopt the SSIM loss as the decode loss. However, SSIM loss was unable to achieve desired performance in the preliminary experiment. Instead, we exploit the \textit{mean square error} loss to guide the CP decoder to extract the copyright image from the encoded $\copyright$styled image, and the result shows less distortion in decoded copyright image. Specifically, the decode loss is expressed as follows.
\begin{equation}
\mathcal{L}_{dec} = MSE(\textbf{I}_{\textit{cp}}, \textbf{I}'_{\textit{cp}} ).
\label{eq:dec}
\end{equation}

By co-joint training two networks, the CP encoder learns how to generate optimal defensive perturbation embedded with copyright information for each input image. At the same time, the CP decoder possesses the ability to decode the decoded $\copyright$styled image that may be distorted.

\begin{figure*}[!htbp]
	\centering
	\begin{minipage}{.12\linewidth}
		\centering
		\includegraphics[width =1\linewidth]{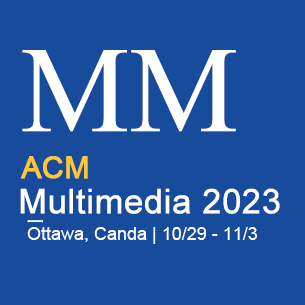}
	\end{minipage}
	\begin{minipage}{.12\linewidth}
		\centering
		\includegraphics[width =1\linewidth]{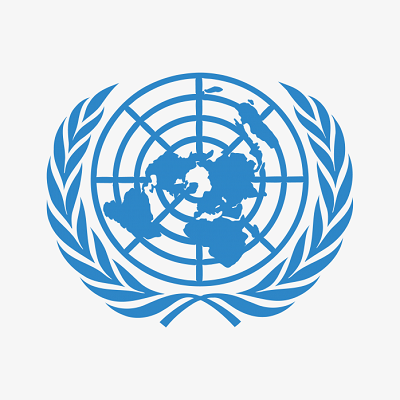}
	\end{minipage}
	\begin{minipage}{.13\linewidth}
		\centering
		\includegraphics[width =1\linewidth]{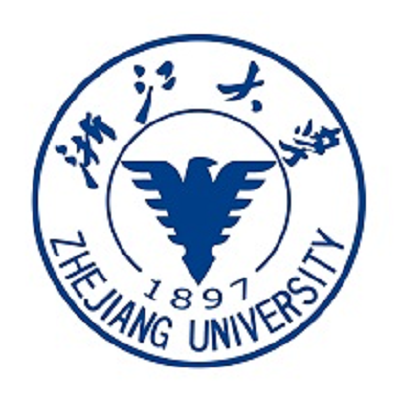}
	\end{minipage}
	\begin{minipage}{.11\linewidth}
		\centering
		\includegraphics[width =1\linewidth]{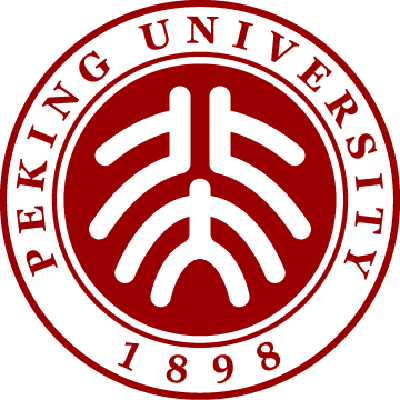}
	\end{minipage}
	\begin{minipage}{.11\linewidth}
		\centering
		\includegraphics[width =1\linewidth]{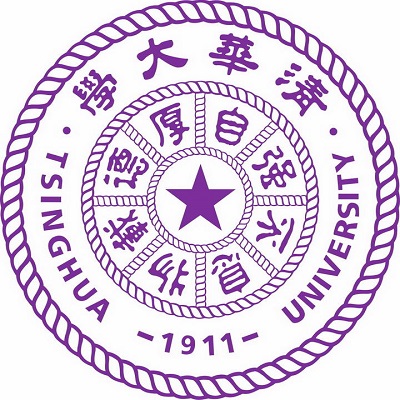}
	\end{minipage}
	\begin{minipage}{.12\linewidth}
		\centering
		\includegraphics[width =1\linewidth]{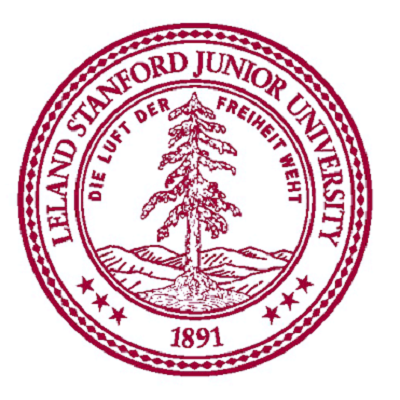}
	\end{minipage}
	\begin{minipage}{.12\linewidth}
		\centering
		\includegraphics[width =1\linewidth]{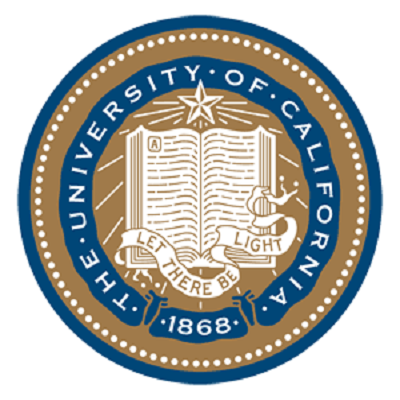}
	\end{minipage}
	\caption{Badges are used for the copyright image.}
	\label{fig:badge}
\end{figure*}

\begin{table*}[ht]
\centering
\footnotesize
\caption{Quantitative results of image quality for source style image and the corresponding $\copyright$encoded image on \textit{summer2winter} dataset.}
\label{tab:result_summer2winter_encoded}
\begin{tabular}{rccccccccccc}
\hline
      			   & \multicolumn{5}{c}{summer2winter}                   &  & \multicolumn{5}{c}{summer2winter\_vangogh}         \\ \cline{2-6} \cline{8-12} 
	 		       & ACM MM     & ZJU        & UCLA     & Stanford & Multi     &  & ACM MM     & ZJU        & UCLA     & Stanford & Multi    \\ \hline
SSIM$\uparrow$     & 0.997983   & 0.9942441  & 0.996534 & 0.997783 & 0.9928219 &  & 0.9988427  & 0.9986982  & 0.996835 & 0.997692 & 0.995783 \\
PSNR$\uparrow$     & 50.5442386 & 46.5713603 & 49.1685  & 50.93422 & 46.812705 &  & 49.5725312 & 48.6787224 & 46.16157 & 47.34184 & 43.95866 \\
LPIPS$\downarrow$  & 0.000041   & 0.000022   & 0.000043 & 0.000028 & 0.000036  &  & 0.000002   & 0.000003   & 0.000002 & 0.000002 & 0.000004 \\ 
FID$\downarrow$    & 0.0811576  & 0.0696325  & 0.116109 & 0.066104 & 0.1534995 &  & 0.0159035  & 0.0215533  & 0.029737 & 0.019584 & 0.055907 \\\hline
\end{tabular}
\end{table*}

\begin{table*}[ht]
\centering
\footnotesize
\caption{Quantitative results of image quality for source copyright image and the corresponding decoded copyright image on \textit{summer2winter} dataset.}
\label{tab:result_summer2winter_decoded}
\begin{tabular}{rccccccccccc}
\hline
				  & \multicolumn{5}{c}{summer2winter}                         &     & \multicolumn{5}{c}{summer2winter\_vangogh}        \\ \cline{2-6} \cline{8-12} 
        		  & ACM MM     & ZJU        & UCLA     & Stanford & Multi      &  & ACM MM     & ZJU        & UCLA     & Stanford & Multi    \\ \hline
SSIM$\uparrow$ 	  & 0.9944434  & 0.9849242  & 0.988773 & 0.991467 & 0.9391595  &  & 0.9988826 & 0.9837815  & 0.978203 & 0.983055 & 0.976727 \\
PSNR$\uparrow$ 	  & 38.7700078 & 35.6527122 & 36.04848 & 35.00411 & 25.6360102 &  & 43.979633 & 34.4954673 & 32.49125 & 29.47688 & 32.00341 \\
LPIPS$\downarrow$ & 0.001134   & 0.002159   & 0.00138  & 0.001042 & 0.008008   &  & 0.000019  & 0.000481   & 0.000694 & 0.00043  & 0.000782 \\ 
FID$\downarrow$   & 17.8243082 & 20.8689009 & 5.962502 & 10.83737 & 101.725174 &  & 1.3438215 & 13.9430032 & 15.69636 & 15.17983 & 37.19172 \\\hline
\end{tabular}
\end{table*}

\section{Experiments}
In this section, we briefly discuss the experiment settings and describe subjective, objective experiments and physical tests detailed to demonstrate the effectiveness of the proposed method.

\subsection{Settings}

\subsubsection{Dataset} To evaluate the effectiveness of the proposed method, we used the two datasets released in \cite{zhu2017unpaired}, including \textit{summer2winter} and \textit{monet2photo}. For the style image, we adopt the Van Gogh starry night, which is used in the previous work \cite{gatys2016image}. Thus, we totally constructed four datasets: two origin datasets (i.e., \textit{summer2winter} and \textit{monet2photo}) and two styled image datasets (i.e., \textit{summer2winter\_vangogh} and \textit{monet2photo\_vangogh}). We adopt the badges shown in Figure \ref{fig:badge} as the copyright image. All images are resize to $400 \times 400$ in the experiment.

\subsubsection{Evaluation Metrics} To quantify the performance of the proposed method, we adopt two traditional subjective measurement metrics: SSIM and PSNR, these two metrics the larger, the better (denoted as $\uparrow$); and two novel network-based metrics: FID \cite{heusel2017gans} and LPIPS \cite{zhang2018unreasonable}, these two metrics the smaller, the better (denoted as $\downarrow$). We calculate the above metrics by adopting the implementation provided in \cite{pytorch2020ignite} except LPIPIS\footnote{https://github.com/S-aiueo32/lpips-pytorch.}. 


\subsubsection{Implementation details} As for the architecture of the CP encoder and CP decoder, we modify the semantic segmentation network BiSeNet \cite{yu2018bisenet} to adapt to our task. Specifically, we use five convolutional layers as the pre-encoder layer to encode the copyright information, which is concatenated with the input image to feed into the CP encoder. As for the CP decoder, we place the STN network in front of the CP decoder to enhance its anti-disturb capability. We adopt the style transfer technique reported in \cite{gatys2016image} to perform style transfer, and construct the styled image in advance to improve the training efficiency. To train the network, we adopt the Adam optimizer with a learning rate of 1e-4. Unless specificity, we adopt the following setting: $\lambda_1$ set to 10, $\lambda_2$ set to 15, batch size set to 20, iteration step is set to 140000, and the region of JEPG compression quality is $[25,100]$, the kernel size of Gaussian Blur is randomly selected from $\left\{3,5,7,9\right\}$. Additionally, we adopt the train tricks (e.g., warmup) reported in \cite{goyal2017accurate} for better convergence. All experiments are conducted on an NVIDIA RTX3090 GPU (24GB) cluster.\footnote{Code will be released after published.}

\subsection{Objective Evaluation}
In this section, we conduct objective experiments to quantitatively and qualitatively demonstrate the effectiveness of the proposed method. Specifically, we train the CP encoder and CP decoder for four single badges (i.e., ACM MM and school badges, including ZJU, Stanford, and UCLA) on each dataset. Moreover, we also train a CP encoder and CP decoder for multiple badges (i.e., seven badges shown in Figure \ref{fig:badge}, indicates as Multi), which means that the CP encoder can encode different badges and the CP decoder can decode different badges from the encoded image. Once all models are trained, we evaluate their performance on the corresponding test datasets.

\begin{table*}[!h]
\centering
\footnotesize
\caption{Quantitative results of image quality for source style image and the corresponding $\copyright$encoded image on \textit{monet2photo} dataset.}
\label{tab:result_monet2photo_encoded}
\begin{tabular}{rccccccccccc}
\hline
                  & \multicolumn{5}{c}{monet2photo}    &  & \multicolumn{5}{c}{monet2photo\_vangogh}         \\ \cline{2-6} \cline{8-12} 
                  & ACM MM     & ZJU        & UCLA     & Stanford   & Multi     &  & ACM MM     & ZJU        & UCLA     & Stanford & Multi    \\ \hline
SSIM$\uparrow$    & 0.9986563  & 0.9982046  & 0.998334 & 0.995662   & 0.9909046 &  & 0.9991622  & 0.9952184  & 0.998542 & 0.99677  & 0.995331  \\
PSNR$\uparrow$    & 50.4973985 & 50.2628374 & 50.62208 & 47.87548   & 45.3106821&  & 50.4836948 & 41.3076127 & 48.81937 & 46.16383 & 44.9004   \\
LPIPS$\downarrow$ & 0.000074   & 0.000044   & 0.000047 & 0.000029   & 0.000053  &  & 0.000002   & 0.000002   & 0.000002 & 0.000002 & 0.000003  \\ 
FID$\downarrow$   & 0.0785865  & 0.0554932  & 0.066212 & 0.053062   & 0.0997368 &  & 0.0116657  & 0.0276228  & 0.018535 & 0.023764 & 0.039467  \\\hline
\end{tabular}
\end{table*}

\begin{table*}[!h]
\centering
\footnotesize
\caption{Quantitative results of image quality for source copyright image and the corresponding decoded copyright image on \textit{monet2photo} dataset.}
\label{tab:result_monet2photo_decoded}
\begin{tabular}{rccccccccccc}
\hline
      			  & \multicolumn{5}{c}{monet2photo}                    &  & \multicolumn{5}{c}{monet2photo\_vangogh}         \\ \cline{2-6} \cline{8-12} 
			      & ACM MM     & ZJU        & UCLA     & Stanford & Multi      &  & ACM MM     & ZJU        & UCLA     & Stanford & Multi    \\ \hline
SSIM$\uparrow$    & 0.9957081  & 0.9852319  & 0.982711 & 0.979176 & 0.9739568  &  & 0.9956415  & 0.9867714  & 0.99097  & 0.989031 & 0.977626 \\
PSNR$\uparrow$    & 39.8373342 & 35.8873601 & 33.30881 & 29.09244 & 30.9653603 &  & 40.81725   & 37.7287396 & 36.89675 & 32.99557 & 31.64098 \\
LPIPS$\downarrow$ & 0.002235   & 0.004616   & 0.0047   & 0.005565 & 0.008912   &  & 0.000267   & 0.000538   & 0.000319 & 0.000357 & 0.001003 \\ 
FID$\downarrow$   & 13.3114121 & 15.6515618 & 8.974405 & 11.22648 & 39.8414937 &  & 17.5493986 & 25.0955912 & 5.469122 & 13.93252 & 35.73031 \\\hline
\end{tabular}
\end{table*}

\subsubsection{Quantitative results}
The quantitative results of styled images and the corresponding encoded $\copyright$style images are listed in Table \ref{tab:result_summer2winter_encoded}, and of copyright image and the decoded copyright image are listed in Table \ref{tab:result_summer2winter_decoded} for \textit{summer2winter} and \textit{summer2winter\_vangogh} dataset, respectively. As we can observe, we realized the goals of the excellent image quality of styled images embedded with copyright information. Meanwhile, achieving the high image quality of decoded copyright images. Specifically, the average SSIM, PSNR, LPIPS, and FID between styled (clean) images and encoded $\copyright$styled (clean) images on both \textit{summer2winter} and \textit{summer2winter\_vangogh} are 0.99658, 47.6889, 0.06089 and 0.00001578, respectively. Correspondingly, the results on the copyright image and the corresponding decoded copyright image are 0.98055, 33.8653, 24.74985, and 0.00166611, respectively. The evaluation metric results of \textit{monet2photo} and \textit{monet2photo\_vangogh} are described in Tbale \ref{tab:result_monet2photo_encoded} and Tabel \ref{tab:result_monet2photo_decoded}, which exhibit the similarity performance as \textit{summer2winter}.

\subsubsection{Qualitative results}
\begin{figure*}[!htbp]
  \centering
  \includegraphics[width =1.\linewidth]{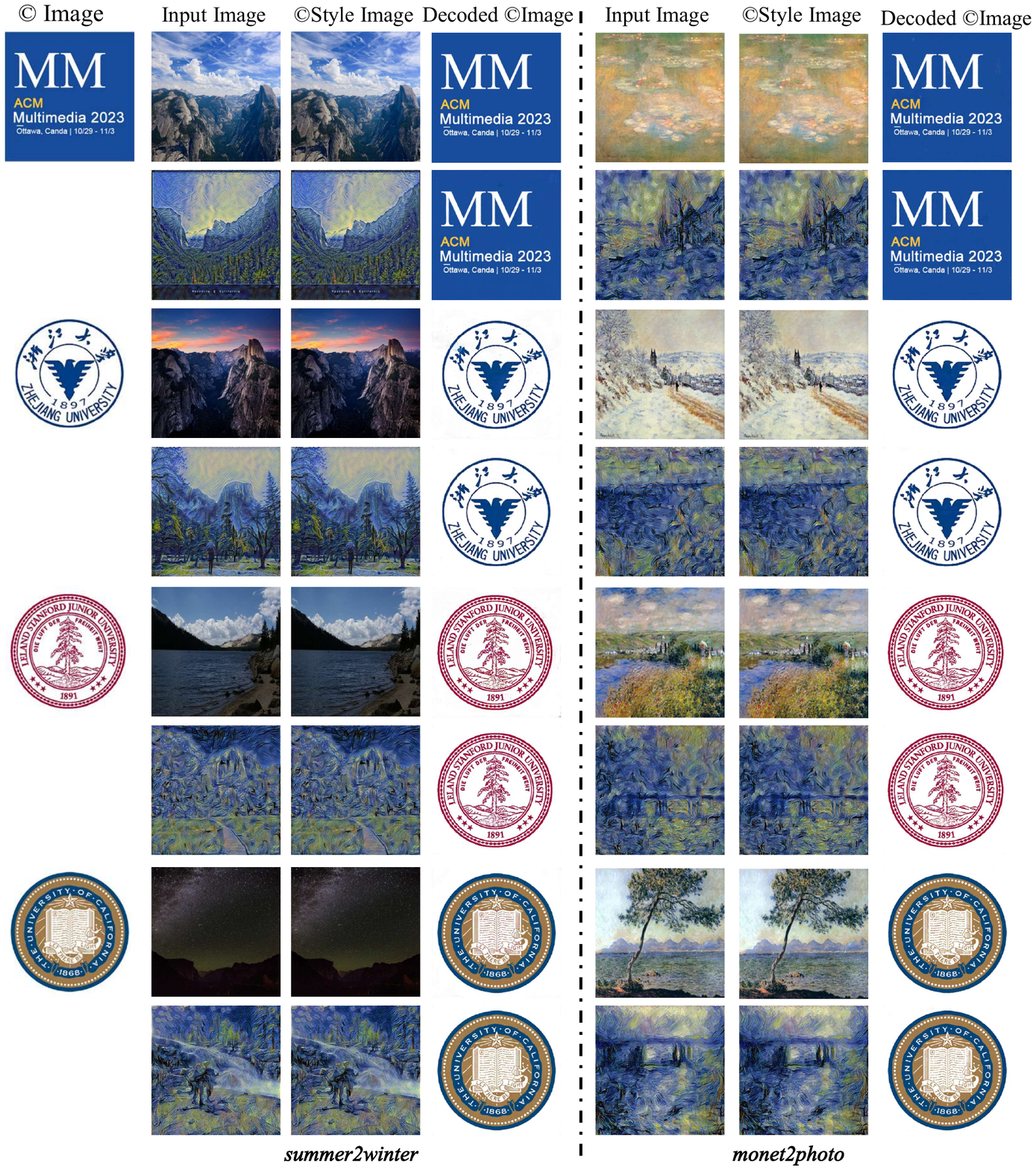}
  \caption{Qualitative results on \textit{summer2winter}, \textit{monet2photo} and their styled transferred variant. From left to right: $\copyright$Image indicates the copyright image; Input Image is the styled image to be protected; $\copyright$Style Image is the styled image encoded with $\copyright$Image; Decoded $\copyright$Image is the decoded copyright image from $\copyright$Style Image.}
  \label{fig:image2show}
\end{figure*}

Apart from the quantitative results, we also provide the qualitative results of different datasets with/without style transfer for Multi badges trained model. Figure \ref{fig:image2show} illustrates visualization results of different embedded badges on different datasets. As we can see, it's hard for humans to distinguish the discrepancy between the input image (styled image) and $\copyright$styled image, and the copyright image and the corresponding decoded copyright image. The result demonstrates that the encoded and decoded copyright images engendered by the proposed method can achieve excellent visual quality.

\subsection{Subjective Evaluation}
The primary goal of our method is to ensure the completeness of the decoded copyright image and simultaneously maintain the image quality after encoding a copyrighted image, which is important in the actual commercial product. Thus, we invite 20 participants (fifteen males and five females) to conduct a subjective experiment to investigate whether the user can perceive the discrepancy between the cleanly styled image and the $\copyright$styled image as well as the source copyright image and the decoded copyright image. Specifically, we devised the two statements (positive/negative) following the Likert scale assessment \cite{likert1932technique}, a widely used approach to scaling responses in survey research, and each statement has four group images that are assessed by five-level scores (1-5). The positive statements require the participant to score the statement "The two displayed images are very similar," where five is "strongly agree," indicating two images are very similar and vice versa. In contrast, the negative statements require the participant to score the statement "The two displayed images are very dissimilar," where five is "strongly disagree," indicating two images are strongly dissimilar and vice versa. To take control experiment, we also provide four group images that are all clean. Therefore, we collected 32 assessments from each participant, 16 for asses clean-encoded clean image (denoted as C-$\copyright$C) pairs and clean-clean (control experiment) image (denoted as C-C) pairs, respectively. The rest 16 assessments were used to assess source copyright and the decoded copyright images pair (denoted as S-$\copyright$S) and source-source image pairs (S-S). To quantity the assessment result, we devised the following metrics
\begin{equation}
	S_{sub} = \frac{1}{P}\sum_{p=1}^{P}(\sum_{i=1}^{N_{pos}} s_{p,i} + \sum_{j=1}^{N_{neg}} (6 - s_{p,j})),
\end{equation}
where $P, N_{pos}, N_{neg}$ indicate the number of participants and the number of image pairs displayed in the positive and negative statements, respectively. In our experiment, $P, N_{pos}, N_{neg}$ are set to 15 (5 for female), 8 and 8. $s_{p, i}$ denotes the score of $i$-th positive question of $p$-th participant. Note that the highest score on the negative question is 5, meaning the image quality is very bad; thus, we convert it by $6-s_{p, j}$ for consistency with the positive one. The higher $S_{sub}$, the better image quality, i.e., the peak $S_{sub}$ is 40 and the neutral $S_{sub}$ is 24. 

Table \ref{tab:subjective_assess} reports the evaluation results. As we can observe, on the one hand, the participant was nearly unable to discriminate the discrepancy between clean images and encoded $\copyright$images. Specifically, the discrepancy between C-C and C-$\copyright$C is 0.55 and 0.2 for males and females, indicating that most participants believe there are no discrepancies between the clean and encoded images. On the other hand, although the female assesses score is lower than males, the score is above neutral and the discrepancy trend is similar. Interestingly, females score in S-$\copyright$S  higher than S-S, meanings that for some participants, the discrepancy between source copyright and decoded copyright image pair are lower than two source copyright image pair. 

\begin{table}[t]
\caption{Subjective assessment results of image quality for clean-clean image pair (C-C) and the clean-$\copyright$style image pair(C-$\copyright$C), source-source copyright image (S-S) and the source-decoded copyright image (S-$\copyright$S) in terms of $S_{sub}$.}
\label{tab:subjective_assess}
\begin{tabular}{ccccc}
\hline
       & C-C        & C-$\copyright$C   & S-S & S-$\copyright$S  \\ \hline
male   & 38.87/40  & 38.33/40         & 38.07/40      & 36.80/40          \\
female & 33.60/40  & 33.40/40         & 33.20/40      & 34.60/40           \\ \hline
\end{tabular}
\end{table}

\begin{figure}[t]
	\centering
	\begin{minipage}{.35\linewidth}
		\centering
		\includegraphics[width =1\linewidth]{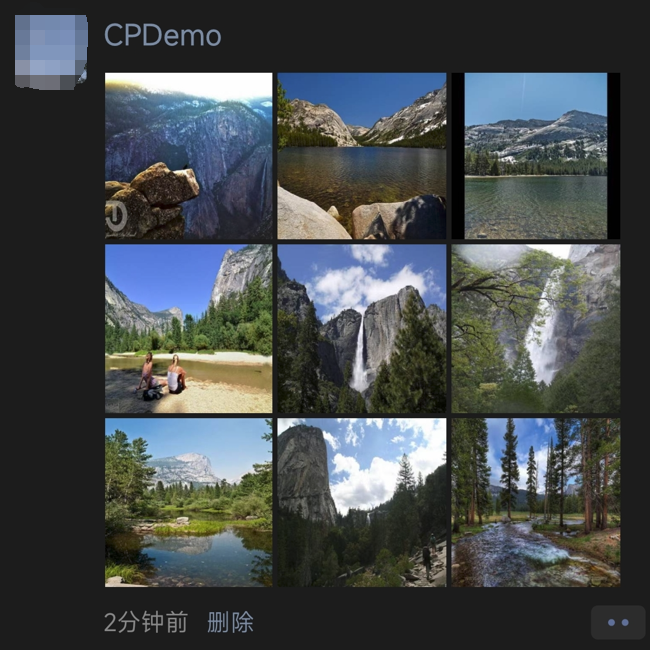}
		\centerline{\footnotesize WeChat}
	\end{minipage}
	\begin{minipage}{.6\linewidth}
		\centering
		\includegraphics[width =1\linewidth]{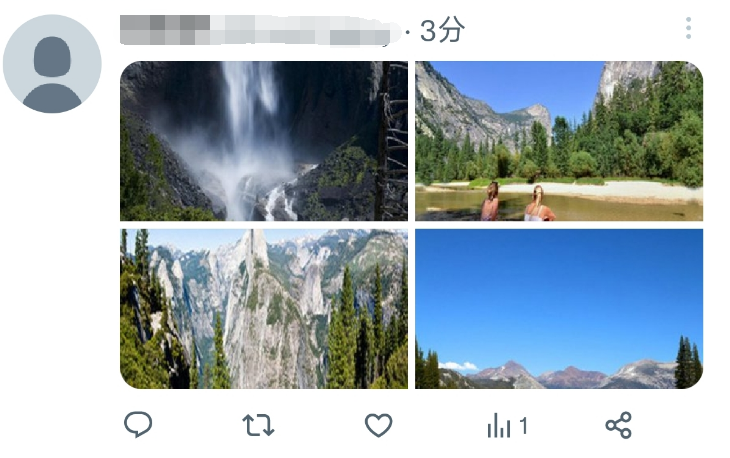}
		\centerline{\footnotesize Twitter}
	\end{minipage}
	\caption{Physical test examples of encoded images posted on social media.}
	\label{fig:social_platform}
\end{figure}

\subsection{Physical Test}
To investigate the effectiveness of the proposed method in real scenarios, such as WeChat and Twitter, the most commonly used social media software, we conduct the physical experiment. Specifically, we first randomly select 50 images, which are embedded with the ACM badge by the CP encoder trained on \textit{summer2winter}. Then, we post the encoded image to WeChat and Twitter (see Figure \ref{fig:social_platform}), respectively. After that, we save the image from the platform hand-by-hand. Consequently, we use the corresponding CP decoder to extract the copyright image (i.e., ACM badge) from them. Table \ref{tab:physical_eval} reports the evaluation metrics between the real copyright and decoded images. We observe that the image posted on social media encounters severe compression: the file size is shrunk to ten percent of the original. Nonetheless, copyright images are decoded corrected. Specifically, on the one hand, the PSNR of WeChat and Twitter are 35.845 and 31.35, which suggests that the image quality is acceptable. On the other hand, the higher SSIM demonstrates the structure of the decoded copyright image is maintained well with the real one. In addition, we observe that although the image detail is lost, the copyright image's content is still recognizable and has competitive image quality. 
\begin{table}[t]
\centering
\caption{Physical evaluation results on WeChat and Twitter.}
\label{tab:physical_eval}
\begin{tabular}{rcc}
\hline
      			  & Wechat   & Twitter  \\ \hline
SSIM$\uparrow$    & 0.991735 & 0.981342 \\
PSNR$\uparrow$    & 36.16807 & 31.37725 \\
LPIPS$\downarrow$ & 0.009821 & 0.019356 \\ 
FID$\downarrow$   & 10.78987 & 18.8384  \\\hline
\end{tabular}
\end{table}

\subsection{Robustness of decoder}
\begin{figure}[t]
	\centering
	\includegraphics[width=.75\linewidth]{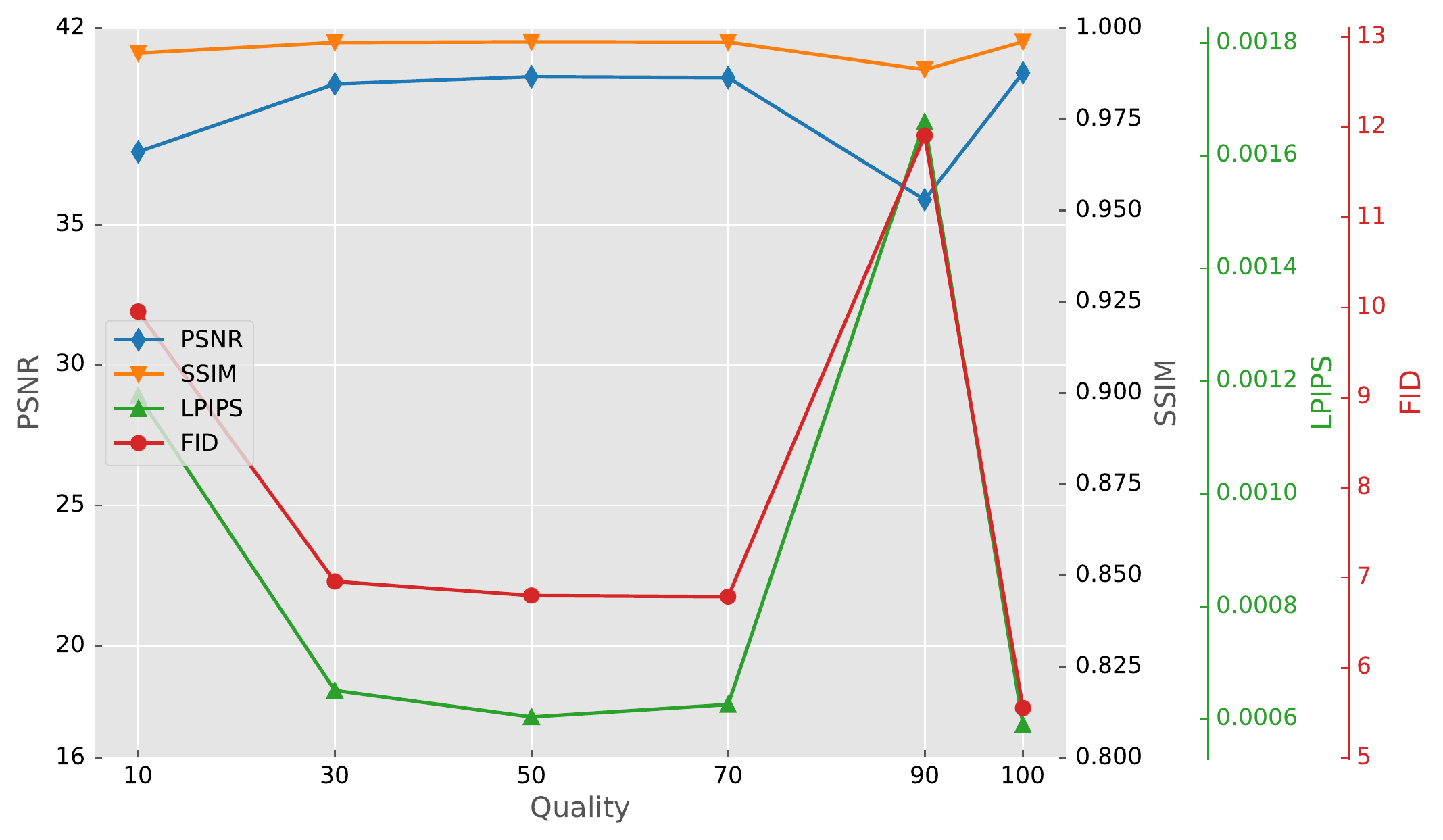}
	\caption{Result of image quality under different JPEG compression qualities.}
	\label{fig:jpeg_compressino}
\end{figure}

In this section, we investigate the robustness of the decoder on different image distortion operations, such as JPEG compression and Gaussian blur. Specifically, we perform JPEG compression with different qualities and Gaussian blur with different kernel sizes on the encoded image. The evaluation results are illustrated in Figure \ref{fig:jpeg_compressino} and Table \ref{tab:gaussian_blur}, respectively. As we can observe, on the one hand, the image quality remains higher despite the JPEG compression quality varying. We speculate the possible reason for the fluctuation of image quality approaching quality 90 is insufficient training caused by random sampling. On the other hand, image quality affected by Gaussian blur exhibits more severe but has nothing to do with the kernel size of Gaussian blur (i.e., the discrepancy among different kernel sizes is small). The possible reason may be the Gaussian blur disturb the distribution of perturbation, which makes the CP decoder unable to decode correctly. Nonetheless, we also find that some distorted examples can be decoded without loss by the CP decoder. However, we expect to increase such training samples may be beneficial to solve this issue. We will investigate this factor in future work. 

\begin{table}[t]
\caption{Result of image quality under different kernel sizes of Gaussian blur.}
\label{tab:gaussian_blur}
\begin{tabular}{ccccc}
\hline
				  & kernel\_3 & kernel\_5 & kernel\_7 & kernel\_9 \\ \hline
SSIM$\uparrow$    & 0.84516   & 0.845912  & 0.845542  & 0.845374  \\
PSNR$\uparrow$    & 22.05886  & 22.1224   & 22.10481  & 22.10274  \\
LPIPS$\downarrow$ & 0.013469  & 0.013417  & 0.013416  & 0.013418  \\ 
FID$\downarrow$   & 179.724   & 178.7791  & 177.8202  & 179.9742  \\\hline
\end{tabular}
\end{table}

\subsection{Discussion}
Someone may be curious about whether encoded images are absolutely equal to input images, while the decoder is tailored to learn how to map the input image to the copyright image. To figure out this question, we use the trained CP decoder to decode clean images. The first and second columns in Table \ref{tab:discuss} report the evaluation image quality results of clean copyright and decoded copyright images. As we can observe, compared with the encoded image, the evaluation metrics of copyright decoded from the clean image showed a significant difference, particularly for FID increased to 244.89, which indicates severe distortion in the image. Moreover, we observe the content in copyright images decoded from clean images is unrecognizable. 

In addition, we observe that some social platforms would inject the visible text watermark into the posted image. Thus, we investigate whether the text watermark impacts encoded images by injecting the “ACM MM 2023” text with font size 18 on images at the randomly selected positions. The third column in Table \ref{tab:discuss} reports the evaluated image quality. As we can observe, the text watermark has a minor impact on the encoded image, which further indicates our method can be used in actual applications. 

Finally, our method can not only be used for protecting DNN-based applications but also be promising in traditional invisible watermarks and steganography. The potential applications of our method can be to protect the high-visual quality demand scenarios, such as copyright of artwork, video, and film.

\begin{table}[t]
\caption{Evaluation results of copyright image decode from different image sources.}
\label{tab:discuss}
\begin{tabular}{rccc}
\hline
      				& Clean Image & Encoded Image & Text Watermark \\ \hline
SSIM$\uparrow$  	& 0.7847514   & 0.9963262     & 0.9953991                    \\
PSNR$\uparrow$  	& 16.205474   & 40.411603     & 39.544393                    \\
LPIPS$\downarrow$   & 0.018105    & 0.0005900     & 0.0007630                    \\
FID$\downarrow$     & 244.89911   & 5.5549556     & 7.1023338                    \\ \hline
\end{tabular}
\end{table}


\section{Conclusion}
With the rapid and wide deployment of AIGC-based applications, an effective and efficient copyright protection method is urgent. In this paper, a plug-and-play copyright protection method based on defensive perturbation is proposed for AIGC-based applications (e.g., style transfer or cartoonish). Specifically, we devise a copyright encoder and decoder to realize the purpose of copyright embedding and extracting, where the copyright information is fused in the defensive perturbation, which can be decoded by the co-trained copyright decoder. To improve the anti-disturb ability, the loss function and robustness module are elaborately devised. Extensive objective, subjective, and physical experiments demonstrate the effectiveness of the proposed method and show its potential usefulness in real scenarios.

{\small
\bibliographystyle{ieee_fullname}
\bibliography{ReviewTemplate}
}

\end{document}